\def\year{2021}\relax
\documentclass[letterpaper]{article} 
\usepackage{aaai21}  
\usepackage{times}  
\usepackage{helvet} 
\usepackage{courier}  
\usepackage[hyphens]{url}  
\usepackage{graphicx} 
\usepackage{natbib}  
\usepackage{caption} 
\usepackage{bbm}
\usepackage{amsmath}
\title{Carbon to Diamond: An Incident Remediation Assistant System From Site Reliability Engineers'  Conversations in Hybrid Cloud 
Operations }
\author {

        Suranjana Samanta,
        Ajay Gupta, 
        Prateeti Mohapatra,
        Amar Prakash Azad\\
}
\affiliations {
    IBM Research India \\
    suransam@in.ibm.com, ajaygupta@in.ibm.com, pramoh01@in.ibm.com,
    amarazad@in.ibm.com
}

\begin{document}

\maketitle

\begin{abstract}
Conversational channels are changing the landscape of hybrid cloud service management. These channels are becoming important avenues for Site Reliability Engineers (SREs) 
to collaboratively work together to resolve an incident or issue. Identifying segmented conversations and extracting key insights or artefacts from them can help engineers to improve the efficiency of the incident remediation process by using information retrieval mechanisms for similar  incidents.  However, it has been empirically observed that due to the semi-formal behavior of such conversations (human language) they are very unique in nature and also contain lot of domain-specific terms. This makes it difficult to use the standard natural language processing frameworks directly, which are popularly used in standard NLP tasks. 
In this paper, we build a framework that taps into the conversational channels and uses various learning methods to (a) understand and extract key artefacts from conversations like diagnostic steps and resolution actions taken, and (b) present an approach to identify past conversations about similar issues. Experimental results on our dataset show the efficacy of our proposed method.
\end{abstract}

\section{Introduction}

In today's hybrid cloud world, understanding why a service fails and what incident remediation steps to perform, with minimal downtime, are extremely challenging tasks. One of the key roles of Site Reliability Engineers (SREs) or IT Operations engineers is to support mission-critical applications and keep the services running. An SRE's understanding of an issue depends on her ability to correctly understand and diagnose the problems from the symptoms and take the best possible action(s) immediately to resolve the issue.

A typical issue management process begins when a critical issue is identified by monitoring systems or the reported explicitly. The SREs analyse the IT operational data, which include error logs, time-series metrics, and ticket records to decide the most effective remediation actions to undertake. Once the issue is resolved, SREs open a new ticket document to record the resolution steps taken to resolve the problem carried out during the investigation. It has been observed that under time and other work pressures, SREs do not create the ticket documents carefully, due to which these documents do not always capture the subtle and key information used and knowledge gained while resolving the issue.
In recent times, SREs' discussions regarding an issue have shifted to collaboration channels, for example, conversational channels \cite{calefato2020case}, \cite{slackAgile19}.
These conversational channels contain a rich set of information exchange regarding the issue at hand, which helps in finding the resolution of the reported issue. 

This paper argues that extracting and storing useful past knowledge from such conversations from the collaboration channels can assist SREs to diagnose the root causes and zeroing to the specific remediation action. This paper proposes a method to learn from SRE conversations, to enable rapid issue resolution and build a rich database of issues, named \textit{Issue Total View}. One of the ways by which mining of such knowledge can significantly reduce the closure time of incidents is by using information retrieval (IR) mechanisms that matches the past conversations to a current issue. While many research efforts have focused on the problem of general issue similarity and IR, these approaches do not generalize well to the IT operations domain, where SRE expertise is often required to determine semantic similarities. We observed that conversational data related to IT operations have their own set of challenges: (1) the presence of hybrid-cloud specific terms, which are unknown to the available pre-trained language models popularly used for various NLP tasks and (2) the semi-formal behaviour of these conversations, which makes it important to identify the important parts of the chat and concentrate on those for further processing.


In this paper, we propose a novel framework for (i) building \textit{Issue total view} database, which contains the diagnostic steps, corresponding actions and the resolutions taken to solve the issues and creates a semi-structured way of representing the conversations for the ease of SREs for future reference and (ii) retrieve past conversations which discuss about a particular query issue, which can help the SREs to check the past resolution steps taken for that particular issue. 



\section{Proposed Method}

\begin{figure*}
  \centering
  \includegraphics[scale=0.4]{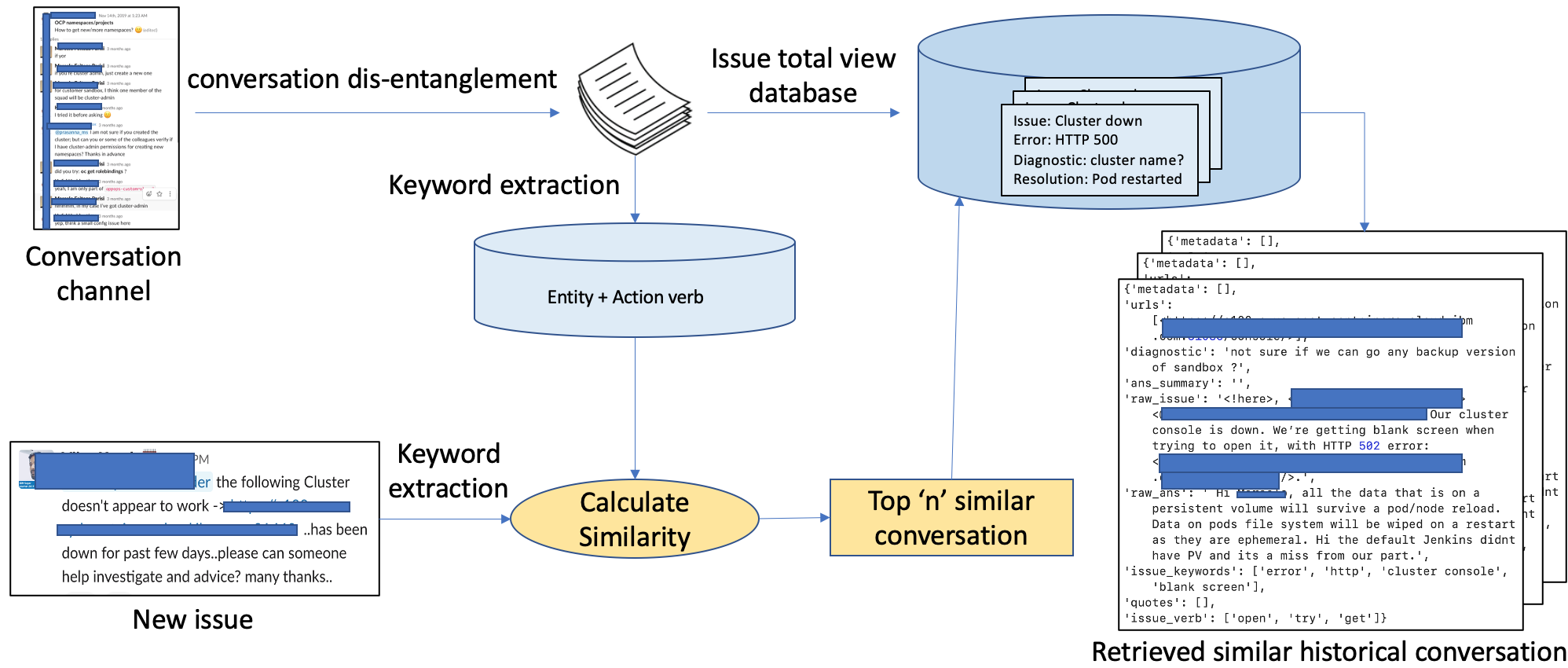}
  \caption{The framework of our proposed approach. Sensitive information have been blocked.}
  \label{fig:flowchart}
\end{figure*}

Figure \ref{fig:flowchart} describes the framework of our proposed method. The input to the framework is the conversation data obtained from a collaboration channel. This data contains all the conversations together, which need to be dis-entangled. The Conversation dis-entanglement module aims to segment each of the conversations from the conversation data. Next, we build the Issue Total View database by extracting the important artefacts from each of the conversations separately. A semi-structured format of the conversation helps SREs to focus on the important sections of the conversation quickly. Finally, we use the proposed similar issue retrieval module to help SREs search through similar contents from the database. We describe each of these modules, in details, in the following sub-sections.

\subsection{Conversation Disentanglement}

Conversation data from collaboration channels consists of all message exchanges among SREs. These messages consist of several different conversation threads. The conversation dis-entanglement module extracts messages which are part of the same thread for further analysis. Lowe et.al. \cite{LowePSCLP17, LowePSP15} proposed a heuristic approach based on time-difference and direct message to extract threads from the Ubuntu corpus. Conversations contained in collaboration platforms, that SREs use, are multi-interlocutor in nature, whereas Ubuntu conversation is a two-way (or dyadic) conversation. SRE's collaboration platforms also have a feature that allows participants to discuss in threaded structure and these native threads can be extracted using the channel's meta-features. But since several participants may not use this feature always, messages may also be written outside these threads as well. These messages can be called  ~\textit{contextual messages}. This module identifies all contextual messages and merges them with relevant threads. Our approach extracts all native threads and extracts potential contextual messages before and after the thread. 
Our approach consists of following rules to identify contextual messages:
\begin{enumerate}
\item ~\textit{Temporal window:} We extract a set of messages within a certain temporal window as potential contextual messages. M\textsubscript{cm}.
\item ~\textit{User overlap:} We extract the set of participant users U\textsubscript{t} from the thread. We extract the set of all users U\textsubscript{c} from potential contextual messages. All messages from the set M\textsubscript{cm} written by $U\textsubscript{t} {\cap}  U\textsubscript{c}$ are considered part of the thread, and are merged together to form one conversation.
\end{enumerate}

\subsection{Issue Total view: Conversation Artefact Extraction}
\begin{figure}
  \centering
  \includegraphics[width=\linewidth]{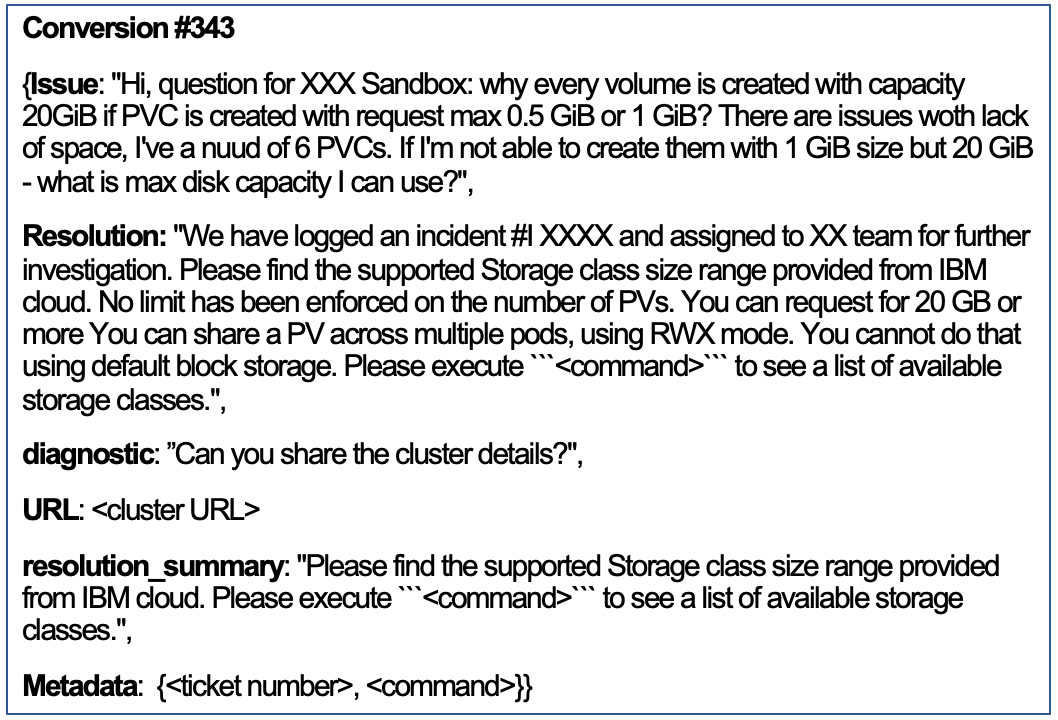}
  \caption{Example of semi-structured form of a conversation as stored in Issue Total View database with extracted artefacts.}
  \label{fig:issue}
\end{figure}

In this sub-section, we build the Issue Total View database using each of the segmented conversations by extracting key artefacts from them. Figure \ref{fig:issue} shows an example of an conversation and its corresponding form in the Issue Total View database. The extracted artefacts give a semi-structured view, helping the engineers to focus on the key sections of the conversation that are relevant and are needed to solve the issue during future references. We present a method for semantic parsing of segmented conversations to extract these artefacts, as explained below.

\subsubsection{Diagnostics Artefact Extraction}


Diagnostic artefact extraction is an important component of information mining for automated incident remediation. Knowing the diagnosis-related utterances can help to understand what investigations were carried out to resolve an issue. In the conversations, we noted that most of the diagnostic utterances are questions or query type statements. Therefore, for this particular conversational data and use-case, the diagnostic artefact extraction mainly identifies question or query utterances from the conversations.
Query utterances, can take both explicit and implicit lexical forms \cite{Mckewon2004,Forsythand2007},
 e.g., explicit: \textit{``Which services are affected ?"}, implicit: \textit{``I was wondering what is the latest impact."}. Furthermore, the queries may also contain informal utterance construct as the conversations are informal in nature. 
To identify the queries, we adopted a semi-supervised approach where we augmented lexical rules along with a simple and effective Naive Bayes classifier. The lexical rules are apt to capture queries containing question words (mainly constituting 5W1H question words, for e.g. who, when, where, what, why, how and presence of '?'), along with a set of other curated verb and adverb based question words (e.g., could, kindly, please). On the other hand, the  Naive Bayes model has been trained on the NPS Chat dataset\footnote{http://faculty.nps.edu/cmartell/NPSChat.htm} to detect the implicit queries which capture the informal query utterances. Our algorithm identifies an utterance as a negative query only when both lexical and Naive Bayes model yield a negative label. The query utterance extraction algorithm achieves an accuracy of about 89.0\% when evaluated on an annotated set created from  our dataset. 




\subsubsection{Resolution Artefact Extraction}

Ayachitula and Khandekar ~\cite{Ayachit20} observed that the verb-noun pair-based approach seems to provide natural and meaningful clusters of IT tickets. 
Following this idea, we extract key entities from a segmented conversation and link them to the required actions. 
We present an approach to establish these links in three phases: (i) extraction of key entities, (ii) extraction of candidate action verbs using a domain-specific dictionary and (ii) link the key entities and action verbs using shallow semantic parser (Semantic Role Labeling) \cite{roth2014composition}.

We use the approach presented in~\cite{mohapatra2018domain} to extract key entities from a conversational utterance. The approach uses various linguistic and non-linguistic features to extract key entities from an utterance. 

We define \textit{action} as a process of performing a change operation by engineers to fix an issue. We consider only those action words which result in a state change of an entity e.g.~\textit{restart}, ~\textit{increase}. The action word dictionary has been populated using an existing Technical Support and Operations corpus from an IT company which consists of change and service request documents. 

We use Semantic Role Labeling (SRL) ~\cite{pradhan2004shallow, Gildea} to extract action-entity links from conversational utterances. SRL  is a shallow semantic parsing task describing who did what to whom, where, when, etc. For each predicate in a sentence, SRL identifies constituents which either play a semantic role (agent, patient, instrument, etc.) numbered from Arg0 to Arg5 or act as an adjunct (location, manner, temporal etc.). Our SRL realization is based on the implementation by Roth ~\cite{roth2014composition} where the annotations are from PropBank~\cite{bonial2010propbank}.  

\begin{figure}
  \includegraphics[width=\linewidth]{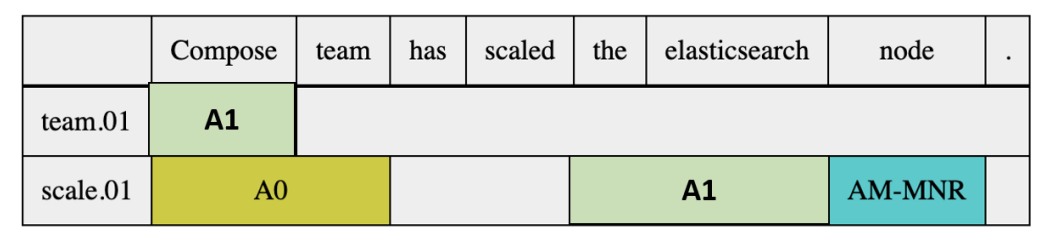}
  \caption{Example of Semantic Role Labeling}
  \label{fig:srl}
\end{figure}

We show in Figure~\ref{fig:srl} the semantic roles with numbered arguments and adjuncts for a sample utterance. An individual sentence can have more than one predicate in it, corresponding to the multiple rows in the output image. Each row in the figure  depicts the label of an argument with respect to a particular predicate. In the example sentence, \textit{``team'', ``scale''} are the two predicates. 
We use a predicate in a sentence, if it's part-of-speech is Verb and if it is presented in the action dictionary. We also explored the relation between semantic role types and ground truth key phrases annotated for each technical document in ~\cite{mohapatra2018domain}, by deriving the distribution for each role type of a key phrase. We obtained the semantic role of each word in the ground truth key phrase from the corresponding sentence to identify the overall frequency distribution of each role in the corresponding dataset. From this experiment, we found that the semantic role \textit{A1} was the most dominating role. Hence, for each predicate in a sentence, we used only the corresponding text that had ‘A1’ as its semantic role. Hence, for the example utterance, we get \textit{(scale, Elasticsearch node)} as the action-entity link. We identify such conversational utterances containing action-entity links as ``resolution\_summary" utterances, a subset of ``resolution" utterances that are all utterances not identified as diagnostic related utternaces.

\subsection{Issue Retrieval Model}
In this part, we describe the method for retrieving similar issues from the Issue Total View database. There have been numerous methods of similarity measures and retrieval techniques in the domain of Natural Language Processing \cite{cer2018universal} \cite{elastic}. However, these techniques are not suitable for our task, where the text has a number of language and domain-specific challenges, such as: \\
\textit{(i) Out-of-Vocabulary (OOV) words} - 
    Most of the existing NLP techniques mask OOV words and limit themselves to a language-specific corpus. However, we want to put emphasis on the domain-specific technical terms, and would also like to find the relations between them. For example, the relatedness  between ``Openshift" and ``Red Hat" should be higher than that between ``OpenShift" and ``Docker". 
    \\
\textit{(ii) Non-issue conversations} - 
    Conversational channels may contain messages pertaining to announcements, informational questions, etc.  In a collaboration channel, it is difficult to identify if an utterance is of issue type or not. Utterances like ``adding users to a service", is more of a change request than an actual issue and hence should be handled differently. 
    It is important to identify the conversations that are related to issues and discard the others.

Next, we present the details of our similar issue retrieval module that tackles the above-mentioned challenges.

\subsubsection{Issue filtering}
We apply the Question Quality Improvement (QQI) \cite{qqi} to categorize the first turn of a  conversation to one of the following classes: ``issue", ``change request" or ``others". QQI helps to identify symptoms present in an utterance by using a deeper understanding of sentences. To detect if the candidate is an issue or not, we use the following rules: (i) If the QQI output detects a symptom, we categorize it into ``issue" class; otherwise (ii) if the utterance contains any action verb (used during artefact extract for building Issue Total View database), we categorize as a ``change request" class; otherwise (iii) we categorize into ``others" class. We proceed with the conversations categorized as ``issue" for further processing as described below.

\subsubsection{Word Embedding}
It is of utmost importance to extract the relations between domain-specific terms like ``Openshift", ``cloudpak" etc., for which we have adapted word-embedding to measure the relationship between them. 
To address the problem of OOV words, we collected documents (troubleshooting guides, manuals, etc) curated from Red Hat\footnote{https://access.redhat.com/documentation/en-us/} and IBM Cloud\footnote{https://cloud.ibm.com/docs} official sites, and use them along with all conversations from the collaboration channel to train a FastText embedding model \cite{fasttext} from scratch. One of the main reasons we preferred FastText over the popular technique of fine-tuning Byte-pair encoding (or its variants lime BERT embedding etc.) is to avoid any bias from the pre-trained weights. For example, the word ``save" is mostly used differently in normal English conversation (aka rescue) and in the domain of computing (aka preserving digital change). Fine-tuning a model on domain-specific data may still have some unwanted bias left from the pre-trained weights. Since our corpus has too many OOV words, training from scratch is expected to give a better embedding model than to fine-tune an existing one. 
FastText model trained on our corpus could identify the word similarities, such as:
\begin{enumerate}
    \item Abbreviation - Abbreviations like ``PV" have words like ``persistent", ``volume", ``persistent volume" within its top ten nearest neighbors. Similarly, ``CLI" has ``command", ``line" as its top ten nearest neighbors in the embedded space. 
    \item Spelling mismatch - FastText could relate words/phrases with different spellings with high similarity. For example, ``Red Hat" is most similar to the word ``RedHat", ``tekon" (an example of wrong spelling) is most similar to the word ``tekton" and ``graffana" is most similar to the word ``grafana". This is a very important property in our case, as in semi-formal conversational settings, users often tend to spell words incorrectly. 
    \item Word relations - FastText beautifully captures the similarity between closely related cloud-specific technical terms. For example: the top three similar words for ``istio" are ``mesh", ``kiali" and ``envoy"; the top three similar words for ``openshift" are ``platform", ``container" and ``cli"; the top three similar words for ``grafana" are ``dashboard", ``alerting" and ``prometheus". Thus any issue with the word ``grafana" will have high similarity with another issue which talks about ``dashboard".
\end{enumerate}

It is to be noted that another popular model of getting word embedding is BERT \cite{devlin-etal-2019-bert}. However, BERT is slower to train as well as to retrieve any word embedding during run-time. As new conversations are included in the training set, it is required to update the embedding model at regular intervals. FastText is preferred as the training is fast and also because it does not need GPU and extra hardware resources for efficient training, unlike BERT. Also, BERT is recommended to be trained for a very large number of epochs (in the order of millions), whereas FastText can learn a good embedding with a significant lesser number of epochs (in the order of hundreds).

\subsubsection{Using Entity and Action Verbs for similarity score}
A text like ``Hello. I have an issue deploying image and it looks like a permission issue." can have high similarity with ``Hello - I have issues with node.js application trying to connect to DB2" because of the common sub-string ``Hello I have an issue". However, these are two different issues, as the first one is about permission, while the second one is about connecting to applications. Also, the two services mentioned in these issues are different. Hence, in order to highlight the importance of the key terms like ``deploying", ``image", ``permission", ``node.js", ``connect", ``DB2", we use the extracted entities and corresponding action verbs, instead of using the entire sentence for similarity measure. We filter out most of the unimportant entities using dependency parsing, where we observe that the key entities have relations of type: clausal modifier of noun (adjectival clause), direct object, the object of a preposition or nominal subject, with any other words in the sentence. We also consider the variants of the action verbs as per WordNet \cite{wordnet}. Each of the sentences are being represented as the list of entities and corresponding action verbs for similarity calculation. Thus, a sentence "Hi, I have not been able to create standard node.js application from the catalog for the last hour" is being represented by the following list of words: ``catalog", ``standard node.js application", ``able", ``create".

\subsubsection{Similarity calculation}
We weigh the entities extracted from the corpus as a function of their Inverse Document Frequency (IDF). If $idf_i$ is the IDF score for the $i^{th}$ entity $e_i$, then the weight is calculated as $w_i = n/(n+idf_i)$, where $n$ is the number of issues present in the database. This function is a non-linear monotonically decreasing function for $0 \leq idf_i \leq 1$ in the range 0 to 1. So any entity with low IDF value has higher weight and is given more importance during similarity calculation. 

\begin{figure}
  \includegraphics[scale=0.4]{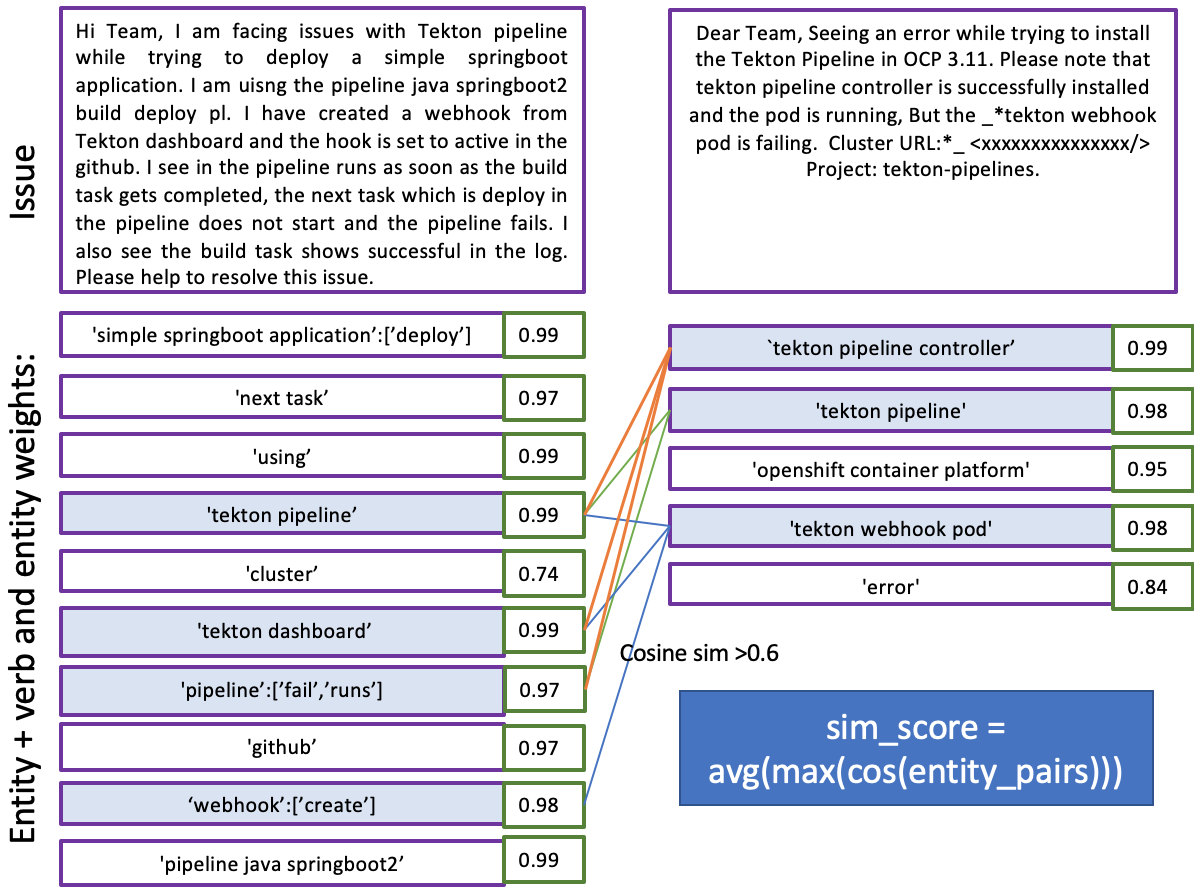}
  \caption{Example of similarity score calculation between two issue texts. The retrieved entities and the corresponding verbs are shown along with their weights, for the two issues. Connection between a pair of entities are shown by a line when the cosine similarity in the embedded space exceeds the value of 0.6. The final similarity score for the two issues shown in the example is 0.72. Sensitive information have been blocked.}
  \label{fig:issue_sim}
\end{figure}

We calculate similarity between two entities $e_i$ and $e_j$ as:

\[
    \delta_{ij}= 
\begin{cases}
    max(w_i,w_j)* J_{ij}& \text{if } J_{ij} > 0.95 \\  & \text{ and } max(w_i,w_j)>0.8\\
    max(w_i,w_j)*(\varepsilon_i \cdot \varepsilon_j),              & \text{otherwise}
\end{cases}
\]
where, $J_{ij}$ is the Jaro distance \cite{string_match} between $e_i$ and $e_j$, and $\varepsilon_i$ denotes $e_i$ in the embedding space. Jaro distance is high ($>$ 0.95) mostly for two words with different spellings. Hence, if $J_{ij} > 0.95$ and $max(w_i,w_j)>0.8$, the entities $e_i$ and $e_j$, are highly distinctive and similar to each other, for which we avoid calculating similarity in the embedding space. This is because the entities are most probably the same word, with spelling variations and their cosine similarity should be ideally 1. In the other scenario, we consider the weighted cosine similarity of the entities in the embedded space, where $\varepsilon_i$ and $\varepsilon_j$ represents the embedding of $e_i$ and $e_j$ respectively. If there is an action verb associated with an entity, we consider the similarity $\delta_{ij}$ only when the same verb (or its variants) is present in both the issue sentence. Let us consider two issues $S_m$ and $S_n$, where there are $l_m$ and $l_n$ number of extracted entities respectively. Then, we represent the similarity between $S_m$ and $S_n$, by the following equation:
\begin{equation}
    issue\_sim(S_m,S_n) = \frac{\sum_i^{l_m} \sum_i^{l_n}{max_j(\delta_{ij})*\mathbbm{1}}}{l_m}
\end{equation}
where $\mathbbm{1}$ is the indicator variable which is 1 when action verb or its variant associated with $e_i$ is also associated with $e_j$ or no action verb is associated with $e_i$. $issue\_sim(S_m,S_n)$ varies in the range 0 to 1 and we retrieve similar issues when the similarity is greater than a pre-defined threshold value. The entire process of similarity score calculation between two sample issues is shown in Figure \ref{fig:issue_sim}. The extracted entities, action verb and weights are shown explicitly.

\section{Results and Analysis}
In this section, we describe our experimental analyses mainly on two dimensions: (1) Efficacy of the conversation dis-entanglement module in terms of identifying segmented conversations (2) Performance in retrieving relevant similar conversations for a given incident using an embedding-based approach over hybrid cloud documents. We evaluated our framework on a collection of $848$ conversations, obtained from a collaboration channel that the operations team uses for its hybrid cloud related operations and support work, gathered over seven months (from October $2019$ to August $2020$). 
On average, the SRE team handles $12$ issues per day in the channel. As mentioned earlier, the channel is also used to broadcast information, and for other informational queries or explicit asks like requesting access to cloud containers, adding members to a cluster, etc.

\subsection{Conversation Disentanglement}
To evaluate the conversation dis-entanglement module, we extracted threads and contextual messages from conversation data. We extracted $189$ threads which are sampled once in $24$ hours time window so that two consecutive threads are separated by at least $24$ hours, for the evaluation purpose. For each thread, we extracted up-to $50$ contextual messages before the first thread message and after the last thread message. These contextual messages are extracted within a two hours window. We extracted $508$ contextual messages for our experiments. For evaluation, we manually annotated contextual messages to identify if messages are part of the thread or not.  The precision and recall of the module is $67\%$ and $83\%$ respectively. As show in Table \ref{tab:conv_seg}, we observed that most of the false positive cases resulted because users who are participants in thread messages are also a participant of contextual messages. Most of these messages are of type \textit{change request access} e.g \textit{@bot Please provide access to $\langle user \rangle$ to cluster}. 



\begin{table}[h]
    \centering
    \begin{tabular}{|c|c|c|c|c|}
   \hline
     conversations & TP & FP & TN & FN  \\ \hline
      189  & 58 & 29 & 409 & 12 \\ \hline
     
    \end{tabular}
    \caption{Performance analysis of the Conversation Dis-entanglement method on our dataset. TP - true positives, FP - false positives, TN - true negatives, FN - False negatives.
    }
    \label{tab:conv_seg}
\end{table}

\subsection{Similar Issue Retrieval}
We compare our proposed approach for retrieving similar issues and their related artefacts with standard state of the art techniques like Elastic search \cite{elastic}, BERT embedding \cite{devlin-etal-2019-bert} and Universal Sentence encoder (USE) \cite{cer2018universal}. While USE finds the entire sentence representation in an embedded space, Elastic search uses TF-IDF based technique to retrieve similar issues from the dataset. Both the methods have been popularly used in similarity score calculation and are robust, which makes them preferable for production. We use BERT embedding, fine-tuned on hybrid-cloud manuals, in the same way as we use FastText embedding. 

\subsubsection{Dataset creation}
From a collection of $848$ conversations, we identified $412$ of them as issue type. Ground truth knowledge of top similar conversations was collected from SREs for $25$ issues. We would like to mention that most of the issues are of similar types and hence the number of distinct issues could not be increased in the annotated set. 

\subsubsection{Experimental evaluation}
The data corpus contains documents (troubleshooting guides, manuals, etc) curated from Red Hat and IBM Cloud official sites. Around $2$ million sentences were extracted from these documents. We used this corpus for training the embedding model, for which we used the BERT Language Model (uncased) and FastText. However, FastText, which was trained from scratch, provided us better word relations. We trained FastText for $300$ epochs to get a 300-dimensional vector representation for each word and considered sub-words from length $3$ to $20$. This enabled us to handle phrases like ``Red Hat", which is a combination of multiple words. For fine-tuning BERT embedding, we used the pre-trained BERT base uncased model and trained it for 300 epochs using the same training corpus, as used in FastText embedding. 

We use Precision ($P@N$), Mean Average Precision ($MAP$), and Accuracy ($A@N$) measures to evaluate the different methods used for similar issue retrieval. Accuracy at $N$ ($A@N$) is the percentage of the queries that have at least one relevant issue retrieved within rank `N'. 
We compare the performance of our approach against three state of the art methods: Elastic search, Universal Sentence Encoder, and BERT. Our proposed method of similar issue retrieval using trained FastText embedding outperforms the other three baseline techniques. We show the results using FastText embedding for various different combinations. Method 1 (FastText-M1) considers the proposed $\delta_{ij}$ for calculating the similarity between a pair of entities, but ignored the action verbs for calculating $issue\_sim$. Method 2 (FastText-M2) shows the result of our proposed method, using the proposed $\delta_{ij}$ and $issue\_sim$. Table \ref{tab:sim_issue} gives the performance of the different methods using the evaluation metrics. As can be seen from the table, the best results comes from using word level embedding as compared to sentence level embedding (using Universal Sentence Encoder).

\begin{table}[h]
    \centering
    \begin{tabular}{|c|c|c|c|c|c|}
    \hline
     Method & P@5 & P@10 & MAP & A@3 & A@5  \\ \hline
       Elastic & 0.21  & 0.28  & 0.28 & 0.40 & 0.42 \\
       USE &  0.10 & 0.11  & 0.13 & 0.21 & 0.21 \\
       BERT &  0.08 & 0.14  & 0.13 & 0.18 & 0.18 \\
       FastText-M1 & \textbf{0.41}  & 0.47  & 0.29 & 0.50 & 0.63 \\
       FastText-M2 & \textbf{0.41}  & \textbf{0.51}  & \textbf{0.31} & \textbf{0.55} & \textbf{0.73} \\ \hline
    \end{tabular}
    \caption{Performance analysis for the task of similar issue retrieval in different experiment setups.}
    \label{tab:sim_issue}
\end{table}

\section{Towards Deployment}
\begin{figure}
  \includegraphics[width=\linewidth]{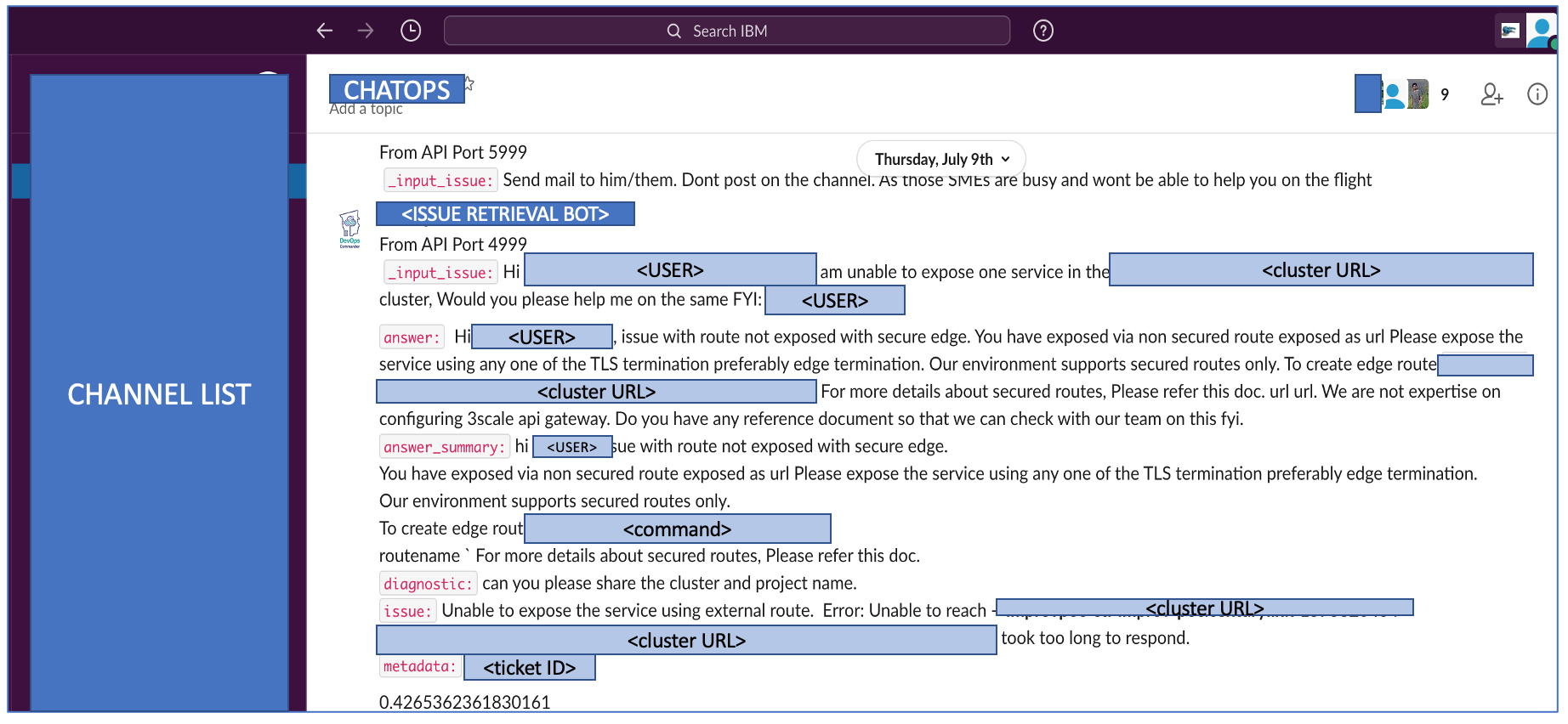}
  \caption{Screenshot of the test collaboration channel using the proposed approach to find similar issues and their artefacts for a given issue (obtained from original SRE discussion collaboration channel) in real-time. Sensitive information has been blocked.}
  \label{fig:deploy}
\end{figure}

The proposed system, Issue Total View, is being deployed in phased manner to assist SRE team for effective and faster remediation of various issues. The objective is to identify similar issues and corresponding artefacts leveraging past conversations, in real-time. 
The integration of the proposed system with an active collaboration channel enables SREs to query issues, evaluate the retrieved results, and provide feedback in real-time. Figure \ref{fig:deploy} shows a snapshot of the test channel when our proposed system was called for one such incident. The ongoing feedback turned out to be highly useful especially in the present conditions where the lack of adequate ground truth inhibits proper performance evaluation. In addition, feedback obtained from SREs can also be used to improve system through various strategies, such as continuous learning, and reinforcement learning. In the first phase, Issue Total View is invoked whenever an incident is reported in the SRE collaboration channel. The results are displayed in a test channel where SREs can provide feedback on the retrieved conversations and artefacts. The feedback obtained from the SREs will be used for further enhancement of the modules.

\section{Conclusion}
In this paper, we propose a novel framework that analyses conversations in a collaboration channel, which is used by 
Site Reliability Engineers (SREs) 
to resolve issues. 
We process the derived 
disentangled conversations and identify key artefacts like issue, diagnostic, and actions or resolutions and use them to create 
Issue Total View 
database. The semi-structured way of representing conversations helps the SREs to focus on the sections of the conversations that are relevant and needed to solve the issue. We train FastText embedding model on documents collected from the hybrid cloud domain, which enables us to learn the domain-specific terms and the relations between them in an effective way. Our framework enhances the ability of the incident remediation process by identifying similar past conversations and their corresponding artefacts. 
Experimental results on our dataset show the efficacy of our proposed framework for this system. In the future, we will further refine the artefact extraction process by identifying other key conversational pieces like actions that resulted in failure or success, actions that resulted in a problem, root cause, stack traces, and error messages. We will also incorporate the feedback of the SREs, for adapting our system in an efficient way. 

\bibliographystyle{aaai21}

\bibliography{main}

\end{document}